\documentclass{article} 
\usepackage{algorithm} 
\usepackage{iclr2025_conference,times}
\usepackage{multirow}

\usepackage{amsmath,amsfonts,bm}

\def\eqref#1{equation~\ref{#1}}

\def\1{\bm{1}}

\DeclareMathAlphabet{\mathsfit}{\encodingdefault}{\sfdefault}{m}{sl}
\SetMathAlphabet{\mathsfit}{bold}{\encodingdefault}{\sfdefault}{bx}{n}

\usepackage{hyperref}
\usepackage{url}
\usepackage{graphicx}
\usepackage{algpseudocode}
\usepackage{float}       

\title{LEWIS (LayEr WIse Sparsity)- A Training Free Guided Model Merging Approach }

\author{Hetarth Chopra, Vidhi Rambhia \& Vikram Adve 
\\
Siebel School of Computing and Data Science\\
University of Illinois at Urbana Champaign\\
Urbana, IL 61820, USA \\
\texttt{\{hetarth2, vidhisr2, vadve\}@illinois.edu} 
}

\iclrfinalcopy 
\begin{document}

\maketitle
\begin{abstract}
As specialized large language models (LLMs) become increasingly prevalent, model merging methods are being used to combine them to create a single multi-task model without requiring any additional data or training. However, these approaches fall short when the objective of merging is to increase the downstream model’s performance on a particular task-specific benchmark. In this work, we propose LEWIS (\textbf{L}ay\textbf{E}r \textbf{WI}se \textbf{S}parsity), a guided model-merging framework that uses activation-based layer importance to dynamically adjust layer-wise task-vector sparsity required for the merge process. LEWIS uses a calibration dataset to prioritize critical layers during the task-vector pruning process required for model merging. This approach guides existing merging methods by preserving essential layer-wise task-specific knowledge while ensuring the merged model performs the best at benchmarks resembling the calibration dataset. Our experiments demonstrate the effectiveness of LEWIS with performance improvements of code instruction-following and math-solving models created through model merging up to 4\% and 11.3\%, respectively, outperforming unguided data-less model merging approaches that use uniform-sparsity. 
\end{abstract}

\section{Introduction}
As specialized large language models (LLMs) fine-tuned for tasks such as math solving or instruction following become more prevalent, efficient model-merging methods have gained critical importance. State-of-the-art techniques like TIES \cite{yadav2024ties}, DARE \cite{yu2024language}, and DeLLA \cite{deep2024della} rely on task vectors \cite{ilharco2022editing}—parameter deltas between a pre-trained model and its fine-tuned variant—to merge models. Although these data-less strategies prune task vectors and fuse them into multi-task models, they often yield only moderate performance across tasks. To address this, recent works such as Model Breadcrumbs \cite{davari2025model}, AdaMerging++ \cite{yang2023adamerging}, and Localize and Stitch \cite{he2024localize} have explored optimizing layer- or parameter-level importance to reduce task interference, but at a higher computational cost. Earlier model-merging methods, including simple averaging \cite{choshen2022fusing}, Fisher-weighted approaches \cite{matena2022merging}, and geometric-based solutions \cite{ainsworth2022git,stoica2023zipit}, often suffer from task interference or disregard crucial details like outlier activations. These outlier activations emerge in large-scale transformers \cite{dettmers2022gpt3} and can be 100 times larger than typical hidden states, making naive pruning detrimental to LLM performance \cite{sun2023simple,wei2024assessing}. Empirical findings also indicate that different fine-tunes exhibit varying layer norms, magnitudes, and angles \cite{jang2025model}, hinting that layer-wise treatment can be beneficial.

In this work, we propose a guided model-merging strategy that augments state-of-the-art methods (e.g., TIES \cite{yadav2024ties} and DARE \cite{yu2024language}) by leveraging insights from Wanda pruning \cite{sun2023simple} and a calibration dataset to fine-tune layer-level task-vector sparsity. Our approach selectively preserves critical task-vector components in the most influential layers during merging, boosting performance on benchmarks that resemble the calibration data. Current merging methods create multi-task models that try to balance individual task performance - achieving an all-rounded performance - we aim to enhance this by adding an additional layer on top.

\section{Methodology}
\label{headings}

\begin{figure}[h]
\begin{center}
\includegraphics[width=1\textwidth]{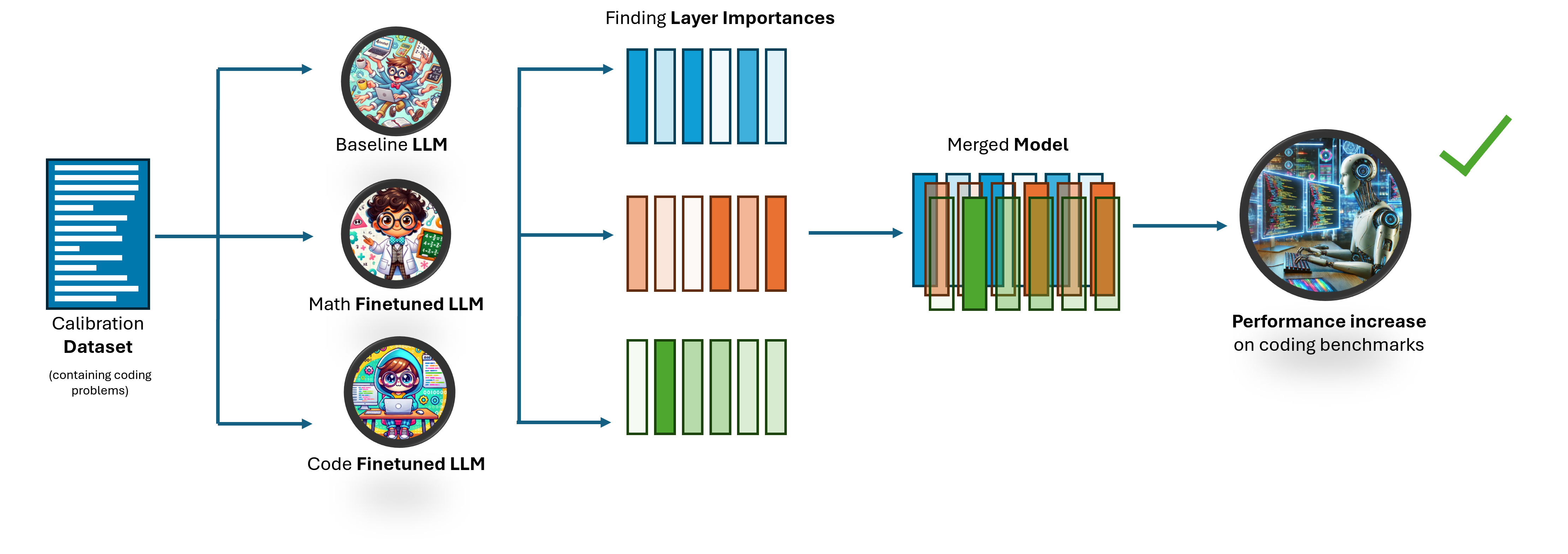}
\end{center}
\caption{Process flow of the LEWIS framework: We show an example of how a calibration dataset (containing coding problems) can be used to compute layer-wise importance for a baseline LLM and it's finetunes, enabling selective ask-vector pruning and merging to perform best on benchmarks containing coding problems.}
\end{figure}


\subsection{Preliminary Notations}
Let an LLM \(f\) be parameterized by \(\theta\) such that \(y = f(x;\theta)\), where $x$ is an input prompt. The pre-trained model has parameters \(\theta_0\). Fine-tuning for task \(T\) yields \(\theta_T = \theta_0 + \Delta\theta_T\), where \(\Delta\theta_T\) captures the changes from \(\theta_0\). Now consider a series of models \(\{\mathcal{M}_p\}_{p=1}^P\), each fine-tuned on a distinct task \(p\). Each model \(\mathcal{M}_p\) is:
\[
\mathcal{M}_p = f(x;\theta_p),\quad \theta_p = \theta_0 + \Delta\theta_p.
\]
Here, \(\Delta\theta_p\) is the task vector (parameter changes for task \(p\)). We can combine fine-tuned models by summing their task vectors. For tasks \(T_1,\dots,T_n\):
\[
\theta_{\text{merged}} 
= \theta_0 + \sum_{i=1}^n \alpha_i \, g(\Delta\theta_{T_i}),
\]
where \(\alpha_i\) are scaling coefficients (responsible for controlling per-model influence in the final merge) and \(g(\cdot)\) is a pruning function (responsible for random/magnitude task-vector pruning function in DARE\cite{yu2024language} and the trimming functionality in TIES \cite{yadav2024ties}). We introduce a calibration set \(\mathcal{D}=\{x_i\}_{i=1}^N\) responsible for guiding the merging process by offering representative data on which we want \(\theta_{\text{merged}}\) to perform effectively on.


\subsection{Lewis: Model Merging using Layer Importance}
\label{subsec:lewis}

Inspired by how Wanda \cite{sun2023simple} works, LEWIS provides a layer importance of an LLM by comparing the activation norms of each layer in the fine-tuned and pre-trained models on a calibration dataset. Layers whose activations deviate more from the pre-trained model are deemed more critical and are pruned less aggressively during the merging process. This selective task-vector pruning ensures that the most critical layers retain higher densities across all fine-tuned models during the model-merging process. The entire methodology can be seen in Algorithm \autoref{alg:lewis}. We first pass the calibration dataset through each fine-tuned model to gather per-layer activations and compute average activation norms, which are compared against the pre-trained baseline (lines~2--12). The resulting differences are normalized and clipped, between empirically tested task-vector sparsity bounds $[\gamma, \epsilon]$, emphasizing layers with significant deviations (lines~14--24). These deviations guide a pruning function that retains parameters crucial for each task (lines~26--30). Finally, weighted task-specific parameter changes are combined to form a single merged model, which is more suited at doing well on a benchmark from which the calibration set \(\mathcal{D}\) is sampled from.

\begin{algorithm}
\caption{Lewis: Model Merging}
\label{alg:lewis}
\begin{algorithmic}[1]
\Require $\theta_0$: Pre-trained parameters, $\{M_p\}_{p=1}^P$: Fine-tuned models with $\{\theta_p\}_{p=1}^P$, $\mathcal{D} = \{x_i\}_{i=1}^N$: Calibration dataset, $[\gamma, \epsilon]$: Sparsity bounds
\Ensure $\theta_{\text{merged}}$: Merged parameters
\For{$p = 1 \dots P$} \For{$x_i \in \mathcal{D}$} \State Get activations $A_{(p, l)}(x_i)$ for all layers $l$ \EndFor \EndFor
\For{$p = 1 \dots P, l$} \State $\text{Norm}_{(p, l)} \gets \frac{1}{N} \sum_{i=1}^N \|A_{(p, l)}(x_i)\|$ \EndFor
\For{$l$} \State $\text{Norm}_{\text{pre-trained}, l} \gets$ Compute similarly \EndFor
\For{$p = 1 \dots P, l$} \State $\Delta A_{(p, l)} \gets |\text{Norm}_{(p, l)} - \text{Norm}_{\text{pre-trained}, l}|$ \EndFor
\State $S \gets \sum_l \Delta A_{(p, l)}$
\For{$p = 1 \dots P, l$} \State $\Delta A_{(p, l)} \gets \text{clip}(\Delta A_{(p, l)} / S, \gamma, \epsilon)$ \EndFor
\For{$p = 1 \dots P, l$} \State $g_{(l, p)} \propto \Delta A_{(p, l)}$ \EndFor \Comment{Higher $\Delta A_{(p,l)}$ = lower pruning rate}
\State $\theta_{\text{merged}} \gets \theta_0 + \sum_p \alpha_p \cdot g(\Delta \theta_p)$
\State \Return $\theta_{\text{merged}}$
\end{algorithmic}
\end{algorithm}

\section{Experiments}
\label{others}


In this section, we evaluate how LEWIS can bootstrap and guide the TIES model merging process in two key scenarios - code instruction-following and math-solving tasks. To achieve this, we use a calibration dataset comprising 15 samples from the training splits of two popular benchmarks for both of these scenarios respectively - MBPP (Most Basic Python Programming) \cite{austin2021program} and GSM8K (Grade School Math 8K). These samples help provide layer-wise importance scores, which inform the design of the pruning function \( g(.) \) used to control task-vector sparsity prior to merging. During the merging process, layers with higher importance scores retain a greater fraction of their task-vectors. Sparsity is constrained within empirically determined task-vector sparsity bounds, \([\gamma, \epsilon]\), to maintain model performance. We compare TIES merging with uniform task-vector sparsity (0.5) for all the layers $l$ having $Q,K,V,O$ and $MLP$ typical of the transformer architecture, as a model merging baseline, with LEWIS. Validation splits from the same benchmarks evaluate the code instruction-following and math-solving capabilities of the merged model. We leverage mergekit \cite{goddard-etal-2024-arcees} for the implementation of our methodology. Both Table \ref{tab:gemma-results} and ~\ref{tab:gsm8k-results} highlight our best-performing results in bold and the second-best results underlined.


\subsection{Scenario 1: Using TIES to create better code instruction following models}
This experiment evaluates the effectiveness of merging Gemma-2b and Gemma-9b \cite{team2024gemma} with their instruction fine-tuned counterparts using LEWIS as a guiding principle for TIES. 
The performance of models was assessed using Pass@1 and Pass@10 scores \cite{chen2021evaluating}. 
Table~\ref{tab:gemma-results} summarizes the results across various task-vector sparsity bounds selected empirically. 
For Gemma-2b LEWIS improves baseline TIES merging using a task-vector sparsity pruning bound of $[\gamma=0.5, \epsilon=0.8]$, by 1.3\% Pass@1 and 4\% Pass@10 scores. Similarly, for Gemma-9b, using LEWIS with a sparsity bound of $[\gamma=0.3, \epsilon=0.8]$ Pass@1 and Pass@10 scores are increased by 1.6\% and 1.8\% respectively. 

\begin{table}[ht!]
\centering
\caption{Performance comparison of Gemma models, code instruction-tuned variants, and merged model across metrics.}
\label{tab:gemma-results}
\begin{tabular}{|l|l|l|c|c|}
\hline
\textbf{Model} & \textbf{Merge Style} & \textbf{Sparsity bounds} & \textbf{Pass@1} & \textbf{Pass@10} \\ \hline
Gemma-2b & N/A & N/A & 0.2746 & 0.3603 \\ \hline
Gemma-2b-Instruction & N/A & N/A & 0.3446 & 0.3829 \\ \hline
\multirow{4}{*}{\shortstack[c]{Gemma-2b + \\ Gemma-2b-Instruction}} 
& Unguided TIES & Uniform 0.5 for all layers & 0.3508 & 0.3850 \\ 
& LEWIS guided TIES & $[\gamma=0.5, \epsilon=1]$ & \underline{0.3536} & \underline{0.3962} \\ 
& LEWIS guided TIES & $[\gamma=0.3, \epsilon=0.8]$ & 0.3456 & 0.3840 \\ 
& LEWIS guided TIES & $[\gamma=0.5, \epsilon=0.8]$ & \textbf{0.3554} & \textbf{0.4004} \\ \hline
Gemma-9b & N/A & N/A & 0.4881 & 0.5718 \\ \hline
Gemma-9b-Instruction & N/A & N/A & 0.5490 & 0.5762 \\ \hline
\multirow{4}{*}{\shortstack[c]{Gemma-9b + \\ Gemma-9b-Instruction}} 
& Unguided TIES & Uniform 0.5 for all layers & 0.5324 & 0.5590 \\ 
& LEWIS guided TIES & $[\gamma=0.5, \epsilon=1]$ & \underline{0.5389} & \underline{0.5706} \\ 
& LEWIS guided TIES & $[\gamma=0.3, \epsilon=0.8]$ & \textbf{0.5408} & \textbf{0.5769} \\ 
& LEWIS guided TIES & $[\gamma=0.5, \epsilon=0.8]$ & 0.5346 & 0.5617 \\ \hline
\end{tabular}
\end{table}

\subsection{Scenario 2: Using TIES to create beter math solving models}

We evaluate the effectiveness of merging LLaMA 3.1 8b \cite{dubey2024llama} with Mathcoder \cite{wang2023mathcoder} leveraging LEWIS to guide TIES merging. We chose Mathcoder, as it has the same architecture as LLaMA 3.1 model, and performs great across different math-solving tasks. 
The performance of the models was evaluated using Flexible-Extract (FE) \footnote[1]{\textbf{Flexible-Extract:} This metric captures numeric answers from text in a broad and adaptable way by using a regex pattern that identifies numbers in diverse formats, such as those with dollar signs, commas, or decimals (e.g., 1,234.56)} and Strict-Match  (SM)
\footnote[2]{\textbf{Strict-Match:} This metric enforces stricter criteria, requiring an exact match to a predefined answer format (e.g., "The answer is -123.45").} metrics.
Table ~\ref{tab:gsm8k-results} summarizes results for baseline models and merged configurations of Llama-3.1-8b and Mathcoder on the GSM8k benchmark. The best performance was achieved with LEWIS-guided sparsity in the range $[\gamma=0.5, \epsilon=0.8]$, outperforming the baseline uniform sparsity TIES method by 11.3\% in FE and 11.2\% in SM. 





\begin{table}[ht!]
\centering
\caption{Comparison of baseline performance for LLaMA 3.1 8b and Mathcoder with results from merged models under different sparsity configurations.}
\label{tab:gsm8k-results}
\begin{tabular}{|l|l|l|c|c|}
\hline
\textbf{Model} & \textbf{Merge Style} & \textbf{Sparsity bounds} & \textbf{FE} & \textbf{SM} \\ \hline
LLaMA 3.1 8b & N/A & N/A & 0.4943 & 0.4928 \\ \hline
Mathcoder & N/A & N/A & 0.6300 & 0.6262 \\ \hline
\multirow{4}{*}{\shortstack[c]{Llama+ \\ Mathcoder}} 
& Unguided TIES & Uniform 0.5 for all layers & 0.5625 & 0.5595 \\ 
& LEWIS guided TIES & $[\gamma=0.5, \epsilon=1]$ & \underline{0.6240} & \underline{0.6217} \\ 
& LEWIS guided TIES & $[\gamma=0.3, \epsilon=0.8]$ & 0.5390 & 0.5390 \\ 
& LEWIS guided TIES & $[\gamma=0.5, \epsilon=0.8]$ & \textbf{0.6262} & \textbf{0.6224} \\ \hline

\end{tabular}
\end{table}


\subsection{Conclusion}

 In this work, we introduced a novel, guided model-merging strategy that builds on top of state-of-the-art merging methods by incorporating layer-wise importance scores. We then leverage calibration data to compute layer-wise activation norms in fine-tuned models, identifying critical parameters that should be preserved during task-vector pruning. This approach mitigates limitations of uniform or purely magnitude-based pruning, thus retaining essential task vectors for improved performance on target benchmarks. Our experiments show its efficiency on both code instruction-following and math-solving tasks: Pass@10 improves by 4\% and 1.8\% for Gemma-2b and 9b, respectively, and FE scores increase by 11.3\% when merging LLaMA 3.1 8b with Mathcoder. These results show LEWIS’s capacity to enhance merged models while preserving critical layer-wise knowledge. Future work will explore automated methods for determining task-vector sparsity bounds and extending this approach to broader task domains, model architectures, and merging strategies.


\bibliography{iclr2025_conference}
\bibliographystyle{iclr2025_conference}

\appendix
\section{Appendix}
This section includes additional ablations that build on these experiments and provide a more
comprehensive view of the merging process. The experiments used NVIDIA GPUs - 4xA100s for the LLM Evaluation;  and 2xRTX5000s for LEWIS and Model Merging process. 

\subsection{Selectively Merging the Top Most Important Layers}

To assess the impact of layer-wise sparsity during the model merging process, we explored a partial merging strategy. This technique involves merging the top $k\%$ most important layers with a density of 1.0, while the remaining layers are merged with a negligible density of 0.1. The results for this experiment, conducted on the \texttt{gemma-2b-it} model, are presented in Table~\ref{tab:partial-merge}

\begin{table}[h]
    \centering
    \caption{Performance comparison of TIES baseline (uniform density) and partial merging strategy with top-$k\%$ important layers as determined by LEWIS.}
    \label{tab:partial-merge}
    \begin{tabular}{|l|c|c|c|}
        \hline
        Merge Strategy & Configuration & Pass@1 & Pass@10 \\
        \hline
        Unguided TIES & Uniform Sparsity 0.5 for all layers & 0.3508 & 0.3850 \\
        \hline
        \multirow{4}{*}{\shortstack[c]{Merge Top-$k\%$ Layers+ \\ as determined by LEWIS }} 
          & $k = 40\%$ & 0.3378 & \textbf{0.3943} \\
                                   & $k = 50\%$ & 0.3419 & \textbf{0.4052} \\
                                   & $k = 60\%$ & 0.3429 & \textbf{0.3973} \\
                                   & $k = 70\%$ & \textbf{0.3556} & \textbf{0.4013} \\
                                   & $k = 80\%$ & \textbf{0.3536} & \textbf{0.4022} \\
        \hline
    \end{tabular}
\end{table}

The results demonstrate that merging based on layer importance, particularly when top-$k\%$ layers are fully preserved, can yield performance improvements. Specifically, for $k = 0.5$, the \texttt{Pass@10} score increased by 5.2\% over the TIES baseline at a uniform density of 0.5. These findings highlight the importance of effectively understanding layer-wise contributions and leveraging them to optimize model merging strategies.

\subsection{Selectively Merging Layers of A Specific Type}

\begin{table}[h]
    \centering
    \caption{Performance of Selected Layer Merge strategy with 100\% density for specific layers and 1\% for others, compared to Unguided TIES.}
    \label{tab:selected-layer-merge}
    \begin{tabular}{|l|c|c|c|}
        \hline
        Merge Strategy & Configuration & Pass@1 & Pass@10 \\
        \hline
        Unguided TIES & Uniform Sparsity 0.5 for all layers & 0.3508 & 0.3850 \\
        \hline
        Selected Layer Merge & Only MLP & 0.3410 & \textbf{0.3927} \\
                            & Only Q & 0.2689 & 0.3335 \\
                            & Only K & 0.2709 & 0.3323 \\
                            & Only V & 0.2480 & 0.3147 \\
                            & Only O & 0.2608 & 0.3247 \\
        \hline
    \end{tabular}
\end{table}

In this experiment, 100\% of the task vectors from a specific layer type (\texttt{Q}, \texttt{K}, \texttt{V}, \texttt{MLP}) were retained during merging, while the remaining layers were pruned to 1\% sparsity. The results for this experiment are summarized in Table~\ref{tab:selected-layer-merge}. This experiment was performed using the \texttt{Gemma-2b + Gemma-2b-Instruction} models, calibrated with 15 samples from the MBPP dataset.

The results reveal that merging layers selectively can significantly influence performance. Notably, retaining the MLP layers at full density while pruning others yields the highest scores for \texttt{Pass@10}. This suggests that MLP layers play a critical role in preserving task-specific knowledge and merit further exploration.
 Conversely, merging only the attention-related layers (\texttt{Q}, \texttt{K}, \texttt{V}, and \texttt{O}) results in comparatively lower performance, underscoring the importance of MLP layers in the model's ability to generalize during the merge process. These insights emphasize the need to account for layer-specific contributions when designing model merging strategies.

\subsection{Scenarios 1 with DARE}

To further validate the performance of layer-wise guided model merging, we repeated the experiments from Scenario 1 (focused on MBPP tasks) using another state-of-the-art merging method, DARE. The performance of Gemma models, including their code instruction-tuned variants and merged configurations, was evaluated using the DARE method with different sparsity configurations. Results, calibrated with 15 MBPP samples, are summarized in Table~\ref{tab:gemma-results}.

\begin{table}[ht!]
\centering
\caption{Performance comparison of Gemma models, instruction-tuned variants, and models merged with DARE across metrics. Bold results indicate the best performance for each metric, while underlined results represent the second-best performance.}
\label{tab:gemma-results-2}
\begin{tabular}{|l|l|l|c|c|}
\hline
\textbf{Model} & \textbf{Merge Style} & \textbf{Sparsity bounds} & \textbf{Pass@1} & \textbf{Pass@10} \\ \hline
Gemma-2b & N/A & N/A & 0.2746 & 0.3603 \\ \hline
Gemma-2b-Instruction & Finetuned & N/A & 0.3446 & 0.3829 \\ \hline
\multirow{4}{*}{\shortstack[c]{Gemma-2b + \\ Gemma-2b-Instruction}} 
& Unguided DARE & Uniform & 0.3321 & 0.3710 \\ 
& LEWIS guided DARE & $[\gamma=0.5, \epsilon=1]$ & \underline{0.3424} & \underline{0.3915} \\ 
& LEWIS guided DARE & $[\gamma=0.3, \epsilon=0.8]$ & 0.3335 & 0.3839 \\ 
& LEWIS guided DARE & $[\gamma=0.5, \epsilon=0.8]$ & \textbf{0.3459} & \textbf{0.3888} \\ \hline
Gemma-9b & N/A & N/A & 0.4881 & 0.5718 \\ \hline
Gemma-9b-Instruction & Finetuned & N/A & 0.5490 & 0.5762 \\ \hline
\multirow{4}{*}{\shortstack[c]{Gemma-9b + \\ Gemma-9b-Instruction}} 
& Unguided DARE & Uniform & \underline{0.5316} & 0.5555 \\ 
& LEWIS guided DARE & $[\gamma=0.5, \epsilon=1]$ & \textbf{0.5377} & \textbf{0.5724} \\ 
& LEWIS guided DARE & $[\gamma=0.3, \epsilon=0.8]$ & 0.4843 & 0.5178 \\ 
& LEWIS guided DARE & $[\gamma=0.5, \epsilon=0.8]$ & 0.5290 & \underline{0.5564} \\ \hline
\end{tabular}
\end{table}

When merging Gemma-2b and Gemma-2b-Instruction, the Unguided DARE method performed worse than the fine-tuned model, yielding 0.3321 (\texttt{Pass@1}) and 0.3710 (\texttt{Pass@10}). However, using LEWIS-guided sparsity, we observed best performance with $[\gamma=0.5, \epsilon=0.8]$, achieving 0.3459 (\texttt{Pass@1}) and 0.3888 (\texttt{Pass@10}), which represents a 4.2\% increase in \texttt{Pass@1} and 4.8\% increase in \texttt{Pass@10} compared to the unguided method. For Gemma-9b, fine-tuning improved its scores to 0.5490 (\texttt{Pass@1}) and 0.5762 (\texttt{Pass@10}). Merging Gemma-9b with its instruction-tuned variant using DARE followed a similar trend with the best performance observed with $[\gamma=0.5, \epsilon=1]$, achieving 0.5377 (\texttt{Pass@1}) and 0.5724 (\texttt{Pass@10}), improving over the unguided DARE method (0.5316, 0.5555) by 1.1\% in \texttt{Pass@1} and 3.0\% in \texttt{Pass@10}.

\subsection{Scenario 2 With DARE}

We also repeated the experiments from Scenario 2 (focused on math-solving tasks) using DARE. Similar to TIES, this experiment was conducted with models calibrated using 15 samples from the GSM8k dataset, and the results are presented in Table~\ref{tab:scenario2-dare}.

\begin{table}[ht!]
\centering
\caption{Comparison of baseline performance for LLaMA 3.1 8b and Mathcoder with results from merged models with DARE under different sparsity configurations. The table highlights the best-performing results shown in bold and the second-best results underlined.}
\label{tab:scenario2-dare}
\begin{tabular}{|l|l|l|c|c|}
\hline
\textbf{Model} & \textbf{Merge Style} & \textbf{Sparsity bounds} & \textbf{FE} & \textbf{SM} \\ \hline
LLaMA 3.1 8b & N/A & N/A & 0.4943 & 0.4928 \\ \hline
Mathcoder & N/A & N/A & 0.6300 & 0.6262 \\ \hline
\multirow{4}{*}{\shortstack[c]{Llama+ \\ Mathcoder}} 
& Unguided DARE & Uniform 0.5 for all layers & 0.0622 & 0.0531 \\ 
& LEWIS guided DARE & $[\gamma=0.5, \epsilon=1]$ & \textbf{0.6240} & \textbf{0.6217} \\ 
& LEWIS guided DARE & $[\gamma=0.3, \epsilon=0.8]$ & \underline{0.5390} & \underline{0.5390} \\ 
& LEWIS guided DARE & $[\gamma=0.5, \epsilon=0.8]$ & 0.3230 & 0.3184 \\ \hline

\end{tabular}
\end{table}

However, incorporating LEWIS-guided sparsity restored performance dramatically. The best performance was observed with $[\gamma=0.5, \epsilon=1]$, reaching 0.6240 (\texttt{FE}) and 0.6217 (\texttt{SE}), a near full recovery to Mathcoder's fine-tuned performance.
The second-best configuration, $[\gamma=0.3, \epsilon=0.8]$, also showed strong results with 0.539 (\texttt{FE}) and 0.539 (\texttt{SM}), significantly outperforming the unguided merging approach. Uniform Pruning with DARE underperformed guided pruning across all configurations, highlighting the importance of structured pruning strategies in preserving task-specific knowledge. These findings reinforce the generalization of LEWIS across different merging methods like TIES and DARE.


\end{document}